# A Multi-Head Attention Soft Random Forest for Interpretable Patient No-Show Prediction

Ninda Nurseha Amalina, Heungjo An[1]

Department of Industrial Engineering, Kumoh National Institute of Technology, Gumi City, South Korea

**Abstract**

Unattended scheduled appointments, defined as patient no-shows, adversely affect both healthcare providers and patients' health, disrupting the continuity of care, operational efficiency, and the efficient allocation of medical resources. Accurate predictive modeling is needed to reduce the impact of no-shows. Although machine learning methods, such as logistic regression, random forest models, and decision trees, are widely used in predicting patient no-shows, they often rely on hard decision splits and static feature importance, limiting their adaptability to specific or complex patient behaviors. To address this limitation, we propose a new hybrid Multi-Head Attention Soft Random Forest (MHASRF) model that integrates attention mechanisms into a random forest model using probabilistic soft splitting instead of hard splitting. The MHASRF model assigns attention weights differently across the trees, enabling attention on specific patient behaviors. The model exhibited 93.72% accuracy, 94.77% specificity, 90.23% precision, 89.38% recall, a 91.54% F1 score and AUC 97.87%, demonstrated high and balance performance across metrics, outperforming decision tree, random forest, logistic regression, and naïve bayes models overall. Furthermore, MHASRF was able to identify key predictors of patient no-shows using two levels of feature importance (tree level and attention mechanism level), offering deeper insights into patient no-show predictors. The proposed model is a robust, adaptable, and interpretable method for predicting patient no-shows that will help healthcare providers in optimizing resources.

**Keywords**

**Patient No-Show, Prediction, Multi-Head Attention, Soft Random Forest, Interpretability**

## 1. Introduction

The absence of patients from scheduled appointments, referred to as no-shows, is a major problem in healthcare systems. Unattended appointments can impact both patients and healthcare providers. For healthcare providers, no-shows lead to increased operational costs, decreased operational efficiency, and reduced clinic revenue. For patients, missed appointments disrupt the continuity of care and waste valuable appointment slots, which in turn increases waiting times for others and contributes to patient dissatisfaction [1][2][3]. Given these multifaceted consequences, accurately predicting patient no-shows has become an essential task in optimizing healthcare service delivery.

To address the challenge of reducing no-shows, hospitals and clinics have implemented a variety of operational strategies, such as automated reminder systems, overbooking policies, and predictive modeling techniques to anticipate whether a patient will attend a scheduled appointment [4][5]. Predictive modeling has emerged as a particularly promising tool because of its ability to handle complex patterns in patient behavior.

[1] Correspondence: heungjo.an@kumoh.ac.kr

Because of the inherently uncertain nature of no-show behavior, machine learning techniques such as logistic regression, naïve Bayes, and random forest techniques have been used frequently in previous research [6][7][8][9]. Other techniques, such as AdaBoost, XGboost, gradient boosting (GB), bagging, and support vector machines (SVM), have also been used to predicting patient no-shows [10][11][12]. Complementing these methods, neural network-based models such as deep neural networks (DNNs), artificial neural networks (ANNs), and multilayer perceptrons (MLPs) have been used to better capture high-dimensional, nonlinear relationships in the data [13][14].

While traditional machine learning models such as logistic regression, random forest, and decision tree offer interpretability [6][7][8], they often fall short in capturing complex feature interactions. In contrast, deep learning models such as ANNs and DNNs are capable of learning intricate patterns but generally lack transparency, which limits their adoption in sensitive domains such as healthcare. To bridge this gap between interpretability and predictive power, recent studies have explored integrating attention mechanisms into neural architectures [15].

Indeed, the attention mechanism has significantly enhanced model performance across various domains by enabling models to focus on relevant input features. However, attention-based models can be prone to overfitting and typically operate as black boxes, limiting their interpretability in critical decision-making applications, especially in healthcare systems. Deep learning black box solutions are not easily adopted by medical staff, even if they have high predictive performance [16]. In contrast, tree-based models such as random forest (RF) offer a more interpretable and robust alternative, particularly well-suited for structured healthcare datasets [17].

Motivated by these contrasting strengths and weaknesses, recent efforts have begun integrating attention mechanisms into RF, with applications. For example, self-attention-enhanced RF models have been proposed for breast cancer classification [18], and attention-augmented RF models have demonstrated promising performance on various datasets, including diabetes [19]. Despite these developments, their application to real-world clinical decision-making problems, particularly for patient no-show prediction, remains relatively unexplored.

To address this gap, the objective of this research is to develop a hybrid model that integrates an attention mechanism into soft random forest (SRF). Specifically, we extend traditional RFs by incorporating soft decision trees (SDT) [20], replacing hard splits with probabilistic soft splits. This enhancement allows for smoother decision boundaries and better handling of uncertainty in no-show behavior.

Moreover, because no-show behavior varies across patients and appointment context, an adaptive mechanism is needed to produce patient appointment-specific predictions while maintaining interpretability. Accordingly, we introduce a trainable attention mechanism and a tree reliability parameter to enhance the SRF model. This enhancement enables instance-specific attention weighting over trees, combining the interpretability and robustness of SRF with the adaptive feature focusing capabilities of attention. Our proposed model offers personalized no-show predictions with higher accuracy and greater relevance to real-world healthcare applications.

Another key contribution of this study is the feature importance analysis at both tree and attention levels. While traditional RF models derive feature importance solely from tree splits [21], this dual perspective provides a more comprehensive understanding of the factors driving patient no-show behavior. This improved understanding will help medical staff in making informed scheduling decisions.

By integrating attention mechanisms with SRF, our proposed model aims to balance accuracy, interpretability, and robustness in patient no-show prediction. This study introduces a novel approach to attention-based modeling within probabilistic RF frameworks, offering new insights into improving appointment scheduling and operational efficiency in healthcare systems.

The remainder of this paper is organized as follows. Section 2 reviews the related literature. Section 3 describes the theoretical foundation, proposed framework, data preprocessing, and model architecture. Section 4 presents the experimental setup, results, and analysis. Finally, Section 5 concludes the study and discusses future work.

## 2. Literature Review

The problem of no-shows has been a concern for healthcare research since the early 1980s [22]. Research on patient no-shows has evolved from significant studies in logistic regression [23] to past statistical and probabilistic approaches. As computational advancements emerged, recent researchers have sought to use the power of machine learning and other related techniques to solve the problem of patient no-show. The draw for this research is to optimize medical services by reducing waiting times and minimizing resource wastage, ultimately improving the efficiency of healthcare providers. The growing list of hospital patients places a demand for such improved service in increasing "smart" digital facilities.

To address patient no-show limitations, researchers have explored various methodologies, from probabilistic techniques such as naïve Bayes and Bayesian networks to machine learning methods such as random forest (RF), ANN, and k-nearest neighbor (k-NN) methods [8][24]. Studies [8][24] have demonstrated that decision trees achieving 94.5% accuracy outperform naïve Bayes with an accuracy of 85%, and both collectively outperform logistic regression approaches. These decision trees have been applied to 33,329 dermatology and 21,050 pneumonology samples to develop a system aimed at reducing hospital appointment no-shows [10].

Some studies have solely harnessed regression-based methods as opposed to a machine learning approach. For instance, [25] employed logistic regression to develop 24 unique models for predicting veterans' attendance at healthcare appointments, identifying certain key predictors, such as previous attendance history and the age of the patient, as critical determinants of no-show inquiries. A similar reliance on logistic regression was used in [6]: applied in a bariatric clinic to improve its management, identified additional predictors, such as the distance from home to the clinic and the number of previous appointments. Furthermore, the model in [6] was less impacted by highly personal details, such as age group and gender. A central focus of these studies was the selected predictors of patient no-shows. Preeminent among the common features that were captured for these studies were patient demographics, past no-show history, booking times, and appointment schedules. Additionally, probabilistic approaches have been explored to develop effective strategies for overbooking scenarios. For instance, [26] utilized an integration of the elastic net (EN) variable selection method with the probabilistic Bayesian belief network (BBN). EN was used to establish a feature selection procedure, which was complemented by the predictive functionality of the BBN network. Unlike the naïve Bayes method [27], BNN offered the researchers the ability to investigate the interrelationship between the predictors for no-shows.

Further expanding on past research, [28] sought to solve the problem of patient no-shows by adopting a combination of probabilistic models, such as logistic regression and recent advanced machine learning models. That study utilized logistic regression combined with GB as the advanced machine learning model. GB had the highest accuracy and F1 score (78% and 0.76, respectively) among the techniques considered.

More recently, with the growing influence of artificial intelligence (AI), several studies have applied AI techniques to the problem of patient no-show cases. While some earlier studies focused solely on logistic regression-based methods, more recent studies [11][29] compared techniques such as AdaBoost, SVM, balanced RF, and decision trees separately or in combination against logistic regression-based counterparts.

In light of previous research, [22] contrasted both Hoeffding trees and John's Ripley's algorithm in the modeling and classification of patient appointments. The two algorithms yielded accuracy values of 77.13% and 76.44%, respectively. Similarly, [12] found the superior performance of decision trees over AdaBoost, achieving precision and recall scores of 0.89 and 0.86, respectively, compared to 0.87 and 0.83 for AdaBoost. The authors displayed the outstanding performance of their decision tree models over the alternatives. However, it was also shown that decision trees were highly susceptible to computational complexity limitations.

While prior studies have primarily focused on no-show predictions in individual clinics, [8] addressed this gap by examining the Saudi dental healthcare system. The study applied decision trees, RF, and MLP to forecast patient no-shows, finding that RF outperformed MLP in this specialized setting. Conversely, unlike most studies that focus solely on offline factors, study [30] served as a first in considering a hybrid of offline and online factors for outpatient appointment modeling to predict patient no-show. The study employed various machine learning techniques, including bagging, KNN, boosting, logistic regression, and decision trees for predicting patient no-shows. The work concluded the favorability of the bagging model over the other models for making no-show predictions on online data.

The number of patient no-shows is often found to be lower than the number of attended appointments, leading to biased models. Several studies [11] [30] [31] have employed data balancing to address this issue. In one of these studies [11], RUS boost, balanced RF, balanced bagging, and easy ensemble were used, along with minimizing the weighted average of type I and type II errors. In another study [30], a synthetic minority oversampling technique (SMOTE) was used on the training dataset to obtain a more balanced sample. In another [31], a hybrid of an oversampling minority class and an undersampling majority class technique was used.

Most prior research has focused on machine learning and deep learning architectures to identify patient no-show behavior. Recently, attention mechanisms have been developed to improve classification and regression algorithms by automatically distinguishing the importance of features and weighing the features. Attention mechanisms, first introduced for machine translation in [32], have become crucial elements in neural networks. The foundational concept of attention can be interpreted using a regression model proposed in [33][34]. The use of attention in [32] was a part of a recurrent neural network (RNN)-based encoder–decoder to encode long input sentences. The transformer architecture introduced in [35] eliminates sequential processing and yields highly accurate results for machine translation without recurrent components. Surveys of various attention-based models are available in [15].

RF is a robust alternative to neural networks. Recently, a few studies, including [18], [19], and [36], have begun integrating attention mechanisms into RF. Study [18] applied self-attention with RF for breast cancer classification, while study [19] used an attention mechanism to assign weight to decision trees in RF. Study [36] extended the work of [19], applying attention and self-attention in RF.

Regardless of the significant progress that has been made in past works, none have applied the novel approach of attention mechanisms to handle the hurdle of patient no-show data. This paper aims to fill the gap by introducing a novel approach that adopts a hybrid approach that integrates attention mechanisms with SRF. The proposed model seeks to improve prediction accuracy by leveraging the strengths of each technique, such as interpretability and robustness, while addressing the limitations of previous approaches.

## 3. Materials and Methods

### 3.1 Theoretical Framework

#### 3.1.1 Soft Decision Tree (SDT)

While decision trees are fast and interpretable, they suffer from drawbacks such as overfitting and non-differentiability, making them challenging to be used in gradient-based optimization. SDT addresses these limitations by replacing hard splits with smooth and probabilistic decisions [20]. SDT allows for a continuous gradient, making the model differentiable and suitable for backpropagation during training. The probability decision in each inner node is defined as follows [20]:

$$p_i(x) = \sigma(xw_i + b_i) \tag{1}$$

where filter $w_i$ and bias $b_i$ are learnable parameters in each inner node $i$, $x$ is the input to the model, and $\sigma$ is the sigmoid function.

#### 3.1.2 Random Forest

A random forest (RF) is an ensemble technique that aggregates predictions from multiple decision trees to improve accuracy and robustness [21]. This study incorporates SDT into the RF framework, enabling differentiability for gradient-based optimization. This modified ensemble, referred to as Soft Random Forest (SRF), preserves the robustness of traditional RFs while leveraging the flexibility of SDTs to enhance model adaptability.

#### 3.1.3 Attention Mechanism

The attention mechanism plays a crucial role in enabling models to focus on relevant parts of the input data. It works by assigning weight to the input, allowing the model to assign important features electively. This mechanism has been widely applied in deep learning, particularly in models like the transformer, where self-attention is used to process the input sequence. The attention mechanism consists of three components called query, keys, and values [35].

A previous study of an attention-based random forest (ABRF) [19] introduced an attention mechanism into traditional RF. In ABRF, the attention weight α is assigned based on the instances $x$ that fall into the same leaf node of a decision tree (distance), where closer instances receive more attention and distant instances get less attention. The query, key, and value in ABRF are referred to as $x$, $A_k(x)$, and $B_k(x)$, respectively. Here, the input instance $x$ is query, $A_k(x)$ denotes the average vector of instances that fall into the same leaf node as $x$ in the $k$th tree, and $B_k(x)$ represents the average target value (i.e., the mean of target label vectors $y$) for those instances that fall into the same leaf node as $x$ in the $k$th tree. The distance $D_k$ between query and key is defined as follows:

$$D_k(x, \tau) = softmax\left(-\frac{\|x - A_k(x)\|^2}{2\tau}\right) \tag{2}$$

where $\tau$ is a tuning parameter (temperature) of the softmax function. In the original ABRF, attention weights are computed using several approaches, including a contamination model, a softmax approach, and a combination of the two, via their trainable parameters. ABRF with softmax is formulated as follows:

$$\alpha\,(x, A_k(x), v, z) = softmax\left(-\frac{\|x - A_k(x)\ \circ z\|^2}{2} v_k\right) \quad (3)$$

where $v_k$ is the vector of training parameters, $z$ is the training vector of the feature weights, and the temperature $\tau$ in Equation (2) is incorporated into the training parameter $v_k$.

However, the preliminary study identified the limited performance of the conventional ABRF on our problem. Hence, this study, instead of directly adopting this trainable parameter, proposes a novel trainable parameter $\delta$ to adjust the attention scores dynamically. This parameter is specifically designed to leverage the tree-wise reliability information from the SRF model. By incorporating this new trainable parameter into the distance computation, the model is expected to be more adaptive, dynamically adjusting attention weights based on both the tree reliability and instance distance. More reliable trees (with lower error) receive higher attention, while less reliable trees have a reduced influence. Additionally, this study employs multi-head attention rather than single-head attention to capture diverse patterns across heads and to model complex dependencies and interactions between features, which are crucial in predicting patient no-shows. The details of a new trainable parameter $\delta$ and multi-head attention are described in Section 3.2.3.

### 3.2 Proposed Framework

In this section, we present the proposed framework for predicting patient no-shows. The proposed framework begins with data collection and data preprocessing, followed by model architecture, and concludes with evaluation. The overall framework is illustrated in Figure 1.

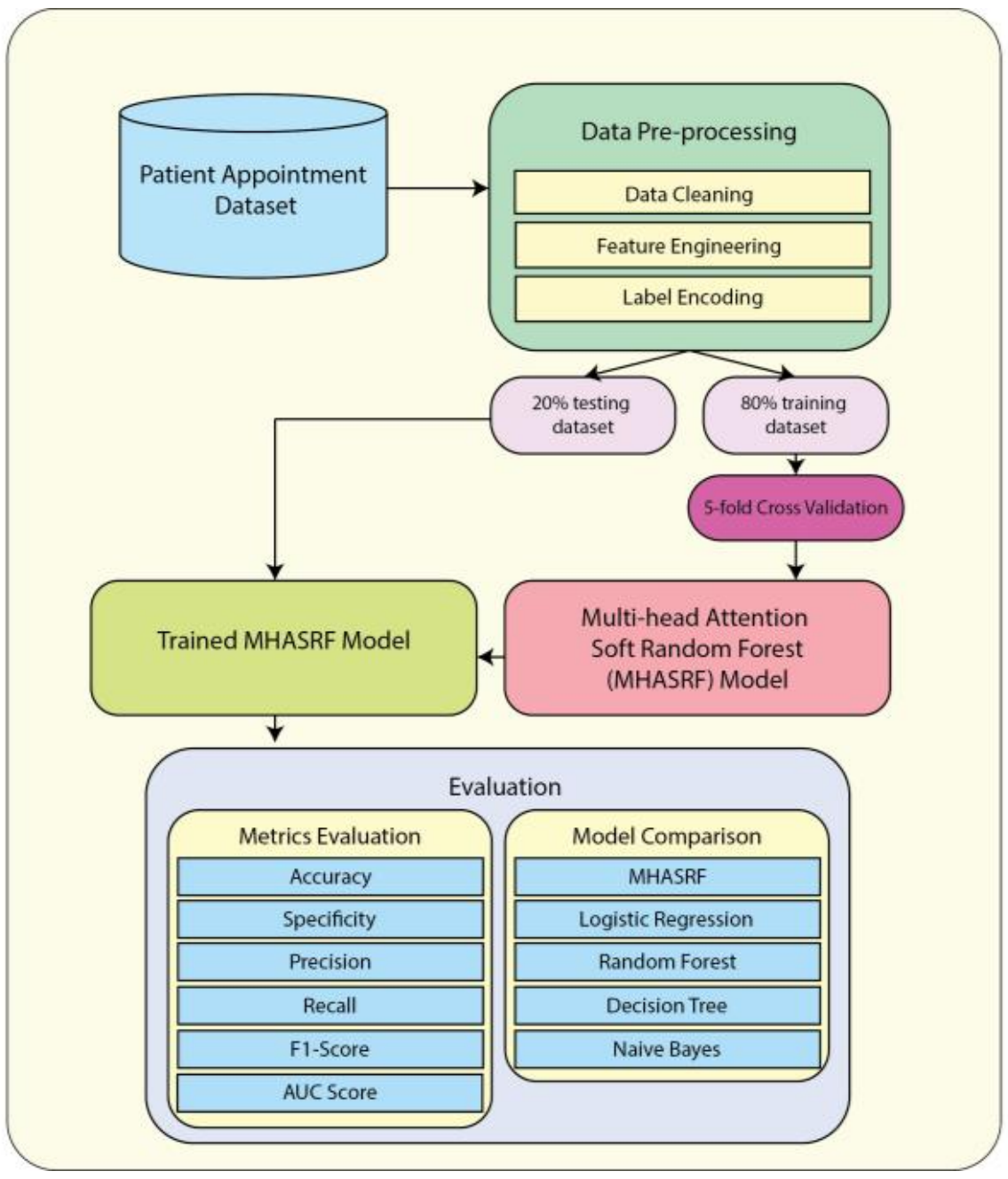

**Figure 1** Proposed Framework

The overall framework begins with preprocessing the patient appointment dataset, which includes data cleaning, feature engineering to enrich the features, and label encoding to convert categorical variables into a numerical format. Once the data are preprocessed, it is split into training and testing sets. The processed data is fed into the predictive model, first processed by SRF. During the training phase of SRF, each tree $k$ generates a mean vector $A_k(x)$ and a mean target value $B_k(x)$ while simultaneously computing a tree error, which is used to estimate the tree reliability parameter $\delta_k$. Following the SRF stage, multi-head attention is applied for each tree and passed through an MLP to enhance the final prediction, and a softmax function is applied as the final step. To ensure robustness, we utilized 5-fold cross-validation during the model training process. In the evaluation phase, we assess the model's performance using several metrics, including accuracy, specificity, precision, recall, and F1 score. The MHASRF model is compared against baseline models such as logistic regression, random forest, decision tree, and naïve Bayes.

### 3.2.1 Data Collection

This study utilized a dataset containing information on patient appointment records from a major healthcare provider in the Middle East. Due to confidentiality agreements, detailed information about the organization cannot be disclosed. The dataset includes appointment records from January 2018 to December 2018, with a total of 157,494 appointments. We only consider patients who either attended (show) or missed the appointment (no-shows), excluding cancelled appointments. Each appointment captured information under the following categories.

- *Patient Characteristics*: patient's age, gender, and language.
- *Appointment Characteristics*: appointment status, type of visit, reason for visit, appointment time, specific healthcare provider, and others.
- *Clinic & Provider*: the type of clinic and name of the physician.

### 3.2.2 Data Preprocessing

To extract useful and relevant information from the dataset, it is essential to preprocess the data before passing it to the model. We conduct data cleaning, feature engineering, and encoding of categorical data to enhance data quality and improve model performance.

- *Data Cleaning*: data cleaning was applied to reduce noise, handle missing data, and remove outliers and irrelevant features.
- *Feature Engineering*: new features were derived from the data. For instance, the percentage of no-shows (% no-show) was calculated based on each patient's appointment history. Additionally, the number of appointments on the same day indicates that a single patient could have multiple appointments within one day. A total of 25 features, including the target variable, were used in this study, as listed in Table 1.
- *Encoding Categorical Values:* the dataset consists primarily of categorical data, such as gender, visit reason, month of the appointment, and visit type. Categorical variables are transformed into numerical representations using label encoding.

After cleaning and standardization of the dataset, the final data used for the model contained 101,532 medical appointments and 24 predictive features. The dataset was split into 80% for training and 20% for testing.

**Table 1** Description of the input features of the model

| Features | Description | Type | Source |
|---|---|---|---|
| **Patient Characteristic** | | | |
| Age | The age of the patient at the time of the appointment | Continues | Raw |
| Language | The preferred language of the patient | Categorical | Raw |
| Gender | The gender of the patient | Categorical | Raw |
| Visit Reason | The reason for the patient's visit | Categorical | Raw |
| **Appointment Characteristic** | | | |
| Visit Type | The type of visit (e.g., procedure visit, consult visit, new visit) | Categorical | Raw |
| Appointment Status | The current status of the appointment (e.g., no-show, show) | Categorical | Raw |
| Time Appointment by Time | The exact time of the appointment | Continues | Raw |
| Time Appointment by Day | The specific day of the appointment is scheduled | Categorical | Raw |
| Time Appointment by Month | The month in which the appointment is scheduled | Categorical | Raw |
| Week of The Month | The week number within the month of the appointment | Categorical | Derived |
| Season | The season during which the appointment takes place (e.g., summer, winter) | Categorical | Derived |
| Number of Visits | Number of visits the patient had before this appointment date/time | Continues | Derived |
| %No-Show | Percentage of the patient's prior appointments that were no-shows | Continues | Derived |
| Number of Appointments on The Same Day | The count of appointments on the same day for the same patient | Continues | Derived |
| **Clinic & Provider** | | | |
| Institute | The healthcare institute where the appointment is scheduled | Categorical | Raw |
| Center Name | The specific center within the institute where the appointment is scheduled | Categorical | Raw |
| Department Name | The department where the patient is visiting (e.g., dentistry, gynecology, urology) | Categorical | Raw |
| Provider Name | The name of the healthcare provider (physician) assigned to the patient | Categorical | Raw |
| **External Factors** | | | |
| Temperature | The temperature at the time of the appointment | Continues | Derived |
| Dew | The dew point (moisture level in the air) during the time of the appointment) | Continues | Derived |
| Humidity | The humidity level at the time of the appointment | Continues | Derived |
| Windspeed | The wind speed at the time of the appointment | Continues | Derived |
| Visibility | The visibility level during the time of the appointment | Continues | Derived |
| Weather Conditions | General description of weather (e.g., raining, cloudy, clear) | Categorical | Derived |
| Air Quality | The description of the air quality index (e.g., good, moderate, unhealthy, hazardous) | Categorical | Derived |

### 3.2.3 Model Architecture

The overall model architecture of MHASRF is illustrated in Figure 2. The model begins with an input layer that receives the preprocessed feature vector $x \in \mathbb{R}^d$, where *d* represents the number of input features. These features are first processed by SRF. During SRF training, each tree *k* maps input *x* to a prediction $B_k(x)$ while computing tree error $C_k$, which determines the tree reliability parameter $\delta_k$. To quantify tree reliability, we define the tree error $C_k$ as the average misclassification rate of tree *k* as given in Equation 4.

$$C_k = \frac{1}{n}\sum_{s=1}^{n} 1(B_k(x_s) \neq y_s) \tag{4}$$

where *n* is the number of samples, $y_s$ is the true value, and $1(\cdot)$ is the indicator function. The tree error $C_k$ and tree reliability parameter $\delta_k$ are both scalars since they each represent a single value per tree. To make $\delta_k(x)$ trainable, we introduce the trainable parameter $\lambda$, leading to the following formulation:

$$\delta_k = \frac{\lambda^H}{C_k} \tag{5}$$

Trees with higher tree reliability parameter $\delta_k$ values have lower error (higher reliability) and contribute more significantly to the final decision, while those with higher error (lower reliability) contribute less.

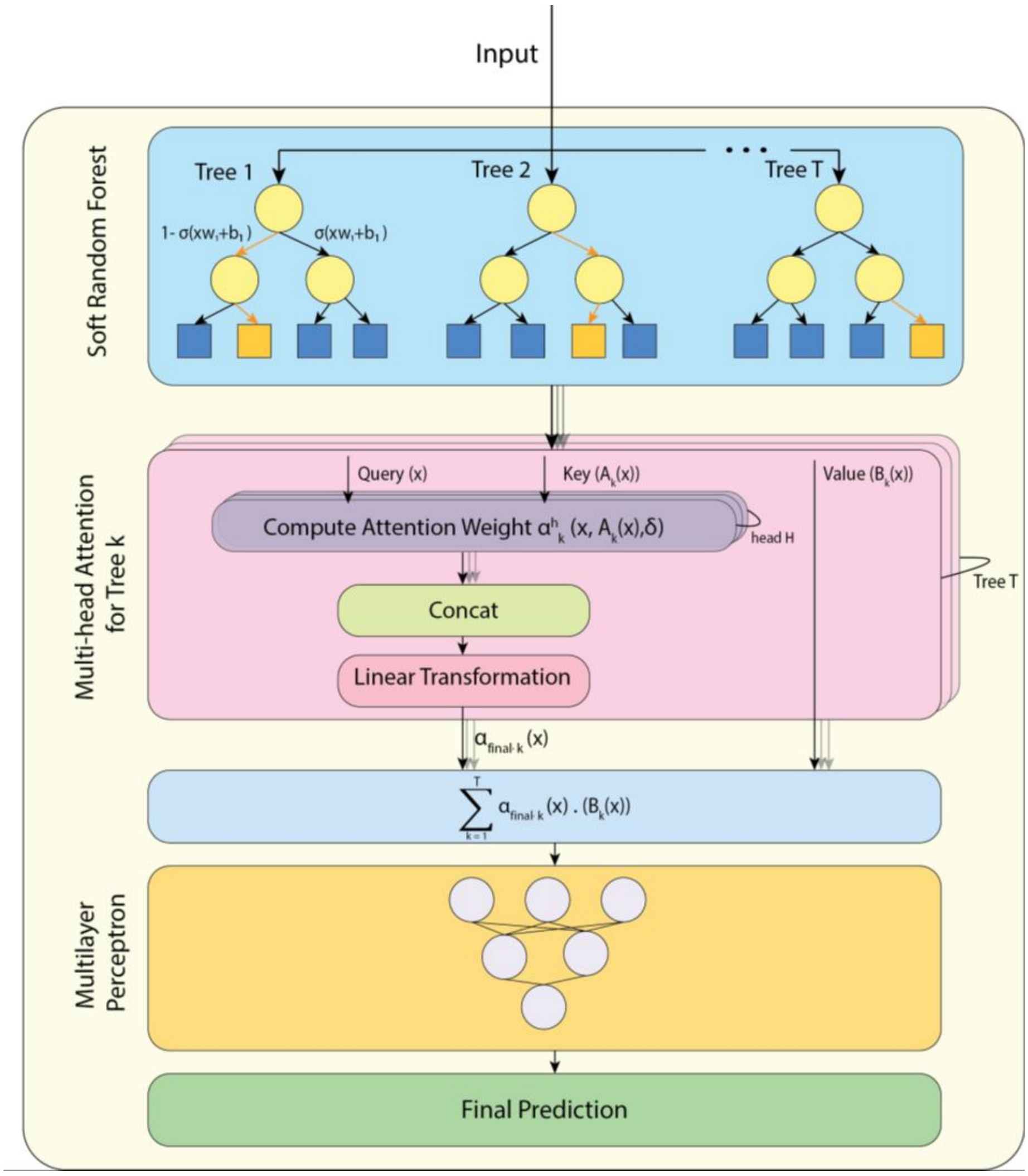

**Figure 2** Model Architecture of MHASRF

Following the SRF, we apply the multi-head attention mechanism, with $x$ as the query, $A_k(x)$ as the key, and $B_k(x)$ as the value. To compute multiple heads attention for each tree $k$, we define the attention weight $\alpha$ for each head as:

$$\alpha_k^h(x, A_k(x), \delta) = softmax\left(\delta_k^h - \frac{\|x - A_k(x)\|^2}{2\tau}\right) \tag{6}$$

where $\tau$ is a temperature parameter. The attention weight $\alpha_k^h$ is a scalar since it represents a single weight per head.

The attention weight $\alpha_k^h$ in Equation 6 combines two important components: the tree reliability parameter $\delta_k^h$ and the distance $\|x - A_k(x)\|^2$. The first term, $\delta_k^h$, reflects how reliable the tree is, with higher values indicating higher reliability. This parameter is derived from the inverse of the tree's error $C_k$, encouraging the model to give more attention to more accurate trees. The second term, $\|x - A_k(x)\|^2$ represents the distance between instance x and the average vector instances that fall into the same leaf node as instance x, reflecting how relevant the tree is for the input. A smaller value of the distance indicates higher relevancy.

By combining the tree reliability $\delta_k^h$ and distance terms, the attention weight $\alpha_k^h$ becomes more adaptive, enabling the model to assign more weight to the trees that are more accurate and relevant to the input. This change is expected to improve overall prediction performance.

After computing the attention weight, we concatenate the attention outputs across all heads into a single vector:

$$\alpha_{concat,k}(x) = [\alpha_k^1, \alpha_k^2, \alpha_k^3, \dots, \alpha_k^H] \tag{7}$$

where $\alpha_{concat,k}(x) \in \mathbb{R}^H$ is an $H$-dimensional vector and $H$ is the total number of attention heads.

To aggregate information across multiple attention heads, we apply a linear transformation using the trainable weight matrix $W_H \in \mathbb{R}^{H \times H}$ that projects the concatenated attention weights for tree k into a new space, allowing the model to learn optimal weight distributions across heads. The final attention weight for tree $k$ that remains an $H$-dimensional vector is defined as follows:

$$\alpha_{final,k}(x) = W_H \cdot \alpha_{concat,k}(x) \tag{8}$$

We then compute a weighted sum of tree predictions, which is passed through a multilayer perceptron (MLP) to enhance the final prediction.

$$y_{final} = MLP\left(\sum_{k=1}^{T} \alpha_{final,k}(x) B_k(x)\right) \tag{9}$$

In the final step of the MHASRF, a softmax function is applied after the MLP to ensure the output is a probability distribution where all class probabilities sum to 1. The final prediction $\hat{y}$ is a vector given by the following:

$$\hat{y} = softmax\,(y_{final}) \tag{10}$$

After computing class probabilities using the softmax function (Eq.10), the final class label is determined by applying a decision rule in which a threshold of 0.5 is used for the binary classification task. The overall training procedure of the MHASRF is summarized in Algorithm 1.

**Algorithm 1**
Multi-head Attention Soft Random Forest

**Input:**
    Dataset $\mathcal{D} = \{(x, y)\}, x \in \mathbb{R}^d, y \in \{0,1\}$
    Number of trees *T*, tree depth *D*, number of attention heads *H*, number of epochs *E*
**Output:**
    Trained MHASRF model
    Global feature importance $I_{final} \in \mathbb{R}^d$
Initialize forest:
  for k = 1 to T:
    $SDT_k$ ← SDT (depth=D, input dim=d, output dim=2)
    add $SDT_k$ to forest
Train SDTs:
  for each $SDT_k$ in forest:
    $\mathcal{D}_k$ ← bootstrap sample of $\mathcal{D}$
    select random subset of features
    for epoch = 1 to *E*:
      compute $B_k(x)$
Compute tree reliability:
  for each $SDT_k$:
    $C_k$ ← average misclassification on $\mathcal{D}$
    $\delta_k \leftarrow \lambda^H / C_k$ , λ is learnable
Multi-head attention:

initialize $\lambda_1 \in \mathbb{R}^H$ and $W \in \mathbb{R}^{H \times H}$
for batch X in $\mathcal{D}$:
  for each tree $SDT_k$:
    compute leaf output $B_k(x)$ and average leaf $A_k(x)$
  for each tree $k$ and head $h$:
    $\alpha_k^h \leftarrow softmax\left(\delta_k^h - \frac{\|x - A_k(x)\|^2}{2\tau}\right)$
  for each tree k:
    $\alpha_{final,k}(x) \leftarrow W_H \cdot \alpha_{concat,k}(x)$
  $y_{final} \leftarrow MLP\left(\sum_{k=1}^{T} \alpha_{final,k}(x) B_k(x)\right)$
  $\hat{y} \leftarrow softmax\left(y_{final}\right)$
Feature importance:
  for each feature $x_j$:
    $I_{tree}(x_j) \leftarrow$ average absolute weight across nodes and trees
    $I_{attention}(x_j) \leftarrow$ attention-weighted tree importance
    $I_{attention}(x_j) \leftarrow$ average the $I_{attention}(x_j)$ and $I_{attention}(x_j)$
Optimize parameters via cross-entropy loss and Adam

### 3.3 Feature Importance

Understanding which features influence patient attendance is essential for reducing no-shows and enhancing the quality of healthcare services. By identifying key factors, hospitals and clinics can take proactive steps to reduce no-shows. Therefore, we compute feature importance using our proposed model by integrating two components: tree-level importance ($I_{tree}$) and attention-level importance ($I_{attention}$).

$I_{tree}$ measures the contribution of each feature within individual trees by quantifying how often a feature is used in decisions at the internal nodes of the tree at the SRF level. SDTs in SRF allow all features to contribute to a node's decision via learnable weights [20]. Accordingly, we define the $I_{tree}$ as the aggregation of the absolute weight across all internal nodes and trees. Formally, $I_{tree}$ for feature $x_j$ in tree $k$ is evaluated as follows:

$$I_{tree,k}(x_j) = \frac{1}{\mathcal{N}_k} \sum_{n=1}^{\mathcal{N}_k} \left| w_{k,n,j} \right| \tag{11}$$

$$I_{tree}(x_j) = \frac{1}{T} \sum_{k}^{T} I_{tree,k}(x_j) \tag{12}$$

where $T$ is the total number of trees, $\mathcal{N}_k$ is the number of internal nodes in the $k$th tree, and $w_{k,n,j}$ is the weight of feature $x_j$ at internal node $n$ in tree $k$. A larger weight indicates a stronger influence of that feature on routing decisions within the tree.

$I_{attention}$ is derived from the final attention weights $\alpha_{final,k}(x)$, which capture the contribution of each tree. A feature is considered more important if it contributes strongly to the decision function of nodes within the trees that receive high attention. $I_{attention}$ is computed as follows:

$$I_{attention}(x_j) = \frac{1}{T} \sum_{k=1}^{T} \alpha_{final,k}(x) \cdot I_{tree,k}(x_j) \tag{13}$$

Both attention levels are scalar. Finally, we combine both global importance measures by averaging both of their contributions. This step ensures that feature importance reflects both its usage in decision trees and the attention mechanism's learned weighting. These importance scores provide valuable insight into the key features influencing patient attendance, empowering hospitals and clinics to implement strategies that reduce no-shows.

### 3.4 Loss Function and Optimization

To optimize the MHASRF model, we employed the cross-entropy loss function, which is commonly used for classification [37]. The loss function is designed to incorporate tree reliability parameters $\delta_k$ and multi-head attention. The loss function is defined as:

$$\mathcal{L} = -\frac{1}{n}\sum_{i=1}^{n}\sum_{c=1}^{C} y_i^c log\hat{y}_i^c \tag{14}$$

where $n$, $C$, $y_i^c$, and $\hat{y}_i^c$ are scalar values that represent the number of training samples, number of classes (no-show and show), true class label, and predicted probability for class $c$, respectively. $\hat{y}_i^c$ is obtained using Equation 10. Model training is performed using the Adam optimizer to update the network parameters.

### 3.5 Evaluation Metrics

Several commonly used evaluation metrics, including accuracy, precision, recall (sensitivity), specificity, and the F1 score, are used to evaluate the performance of a classification model. These metrics are computed based on a confusion matrix consisting of true positive (TP), false positive (FP), false negative (FN), and true negative (TN) values.

TP: This occurs when the model correctly predicts the patient will be a no-show when the actual condition is indeed a no-show.

FP: This occurs when the model incorrectly predicts the patient will be a no-show when the actual condition is that the patient does show up to the appointment.

TN: This occurs when the model correctly predicts the patient will show up to the appointment when the actual condition is indeed that the patient shows up.

FN: This occurs when the model incorrectly predicts the patient will show up when the actual condition is a no-show.

The evaluation metrics are defined as follows:

$$Accuracy = \frac{TP + TN}{TP + TN + FP + FN} \tag{15}$$

$$Precision = \frac{TP}{TP + FP} \tag{16}$$

$$Recall = \frac{TP}{TP + FN} \tag{17}$$

$$Specificity = \frac{TN}{TN + FP} \tag{18}$$

$$F1\ Score = 2 \times \frac{Precision \times Recall}{Precision + Recall} \tag{19}$$

## 4. Experiments and Results

### 4.1 Experimental Setup

Our proposed model was compared against traditional models, such as decision tree, RF, logistic regression, and naïve Bayes classifier models, using accuracy, specificity, precision, recall, and F1 score as evaluation metrics.

We selected these types of models for comparison because they have been commonly used as benchmarks in prior studies [6][7][8][9][27]. The experiments were conducted on a 7th-generation Core i7 CPU with a speed of 2.10 GHz. Python was used with the sci-kit learn and PyTorch libraries to develop the models used in this study.

The dataset was initially divided into 80% for training and 20% for testing. A 5-fold cross validation strategy was applied only to the 80% training set to train and validate the model. Once the model training and cross-validation were completed, the remaining 20% of the data was used exclusively as an independent test set to assess the final performance of the models.

In our experiment, we used 100 trees, a tree depth of three, three attention heads, and a learning rate of 0.01. The selection of these hyperparameters was carefully determined based on multiple preliminary tests in which conditions were varied to produce the best results consistently across different experiments.

### 4.2 Results and Analysis

In this section, we present the experimental results obtained with the MHASRF. We evaluated its predictive performance and compared it with the performance of various traditional methods. The analysis also examines the distribution of attention weights, identifies the key features that significantly influence patient no-show based on feature importance, and includes an ablation study to assess the contribution of model components.

#### 4.2.1 Performance Comparison with Baseline Models

##### 4.2.1.1 Performance on Validation Data (Five-Fold Cross-Validation)

The validation results comparing the MHASRF model with several baseline models are shown in Table 2. A five-fold cross-validation was used to evaluate model performance, where the '±' symbol represents the standard deviation across folds, reflecting the consistency of each model's results. The MHASRF model achieved the highest overall accuracy (93.47% ± 0.28%), supported by strong specificity (94.22% ± 0.56%), precision (88.48% ± 0.95%), recall (91.91% ± 0.81%), F1 score (90.15% ± 0.43%), and an excellent AUC score (97.77% ± 0.11%). These results indicate that MHASRF performs consistently well across all performance metrics and maintains a good balance between identifying no-shows and avoiding false alarms, instances where the model predicts a no-show but the patient actually attends their appointment (FP).

**Table 2** Average Predictive Performance and Standard Deviations Across Models (Five-Fold Cross-Validation)

| Model | Accuracy | Specificity | Precision | Recall | F1 score | AUC score |
|---|---|---|---|---|---|---|
| MHASRF | 93.47% ± 0.28% | 94.22% ± 0.56% | 88.48% ± 0.95% | 91.91% ± 0.81% | 90.15% ± 0.43% | 97.77% ± 0.11% |
| Decision Tree | 93.59% ± 0.21% | 92.55% ± 0.31% | 86.11% ± 0.51% | 95.73% ± 0.08% | 90.66% ± 0.39% | 96.94% ± 0.05% |
| Random Forest | 89.28% ± 0.27% | 97.92% ± 0.13% | 94.30% ± 0.39% | 71.34% ± 0.80% | 81.23% ± 0.60% | 97.64% ± 0.07% |
| Logistic Regression | 91.69% ± 0.13% | 93.93% ± 0.10% | 87.37% ± 0.23% | 87.05% ± 0.41% | 87.21% ± 0.24% | 97.08% ± 0.06% |
| Naïve Bayes | 89.21% ± 0.22% | 90.39% ± 0.16% | 81.31% ± 0.39% | 86.76% ± 0.52% | 83.95% ± 0.39% | 94.32% ± 0.17% |

Among the baseline models, the decision tree achieved the highest recall (95.73% ± 0.08%), meaning it was slightly better at identifying true no-show cases (TP). However, this came at the cost of lower precision (86.11%

± 0.51%) and specificity (92.55% ± 0.31%), suggesting that it also produced more false positives (FP). In contrast, the RF achieved the best precision (94.30% ± 0.39%) and specificity (97.92% ± 0.13%), but its recall dropped significantly (71.34% ± 0.80%), showing that it failed to detect many actual no-shows (TP). The logistic regression and naïve Bayes models performed moderately, but neither offered the same balanced results as MHASRF.

#### 4.2.1.2 Performance on Held-Out Test Data

To further evaluate the generalization ability of the MHASRF model, we assessed its performance on a held-out test set that was not used during training or cross-validation. Results on the held-out test set demonstrate the superiority and stability of the MHASRF model. As shown in Table 3, MHASRF achieved the highest accuracy (93.72%), along with strong specificity (94.77%), precision (89.38%), recall (91.54%), F1-score (90.44%), and the highest AUC (97.87%). These results confirm that MHASRF generalizes well beyond the training data and maintains consistent performance when faced with new cases. Its recall remains high, meaning that most patients who are likely to miss their appointments are correctly identified, which is essential for reducing the operational impact of patient no-shows.

Although the decision tree again achieved slightly higher recall (95.77%), it showed lower precision (85.73%) and specificity (92.34%), indicating that it tends to predict more false positives. Meanwhile, the RF achieved high precision (94.03%) and specificity (97.78%) but low recall (72.88%), which means it would miss many actual no-shows. The logistic regression and naïve Bayes models offered balanced yet weaker results across most metrics. Compared with these models, MHASRF consistently provides the best trade-off between sensitivity and precision, making it the most reliable model for detecting patient no-shows while maintaining a reasonable false alarm rate.

**Table 3** Comparison of Predictive Performance Across Models on Held-Out Test Data

| Model | Accuracy | Specificity | Precision | Recall | F1 score | AUC score |
|---|---|---|---|---|---|---|
| MHASRF | 93.72% | 94.77% | 89.38% | 91.54% | 90.44% | 97.87% |
| Decision Tree | 93.45% | 92.34% | 85.73% | 95.77% | 90.47% | 97.02% |
| Random Forest | 89.69% | 97.78% | 94.03% | 72.88% | 82.11% | 97.71% |
| Logistic Regression | 91.68% | 93.65% | 86.89% | 87.59% | 87.24% | 97.16% |
| Naïve Bayes | 89.40% | 90.40% | 81.39% | 87.30% | 84.24% | 94.61% |

In a healthcare system, a scheduled patient not showing up (i.e., FN) can result in resource wastage, disruption to continuity of care, longer waiting times, and increased operational costs [1][2][3]. Conversely, a patient expected to be a no-show actually attending (i.e., FP) may lead to overbooking or rescheduling, which is disruptive but less critical than an FN. Since the primary goal of no-show prediction is to accurately identify patients who will miss their appointments, our model prioritizes recall (sensitivity) over specificity. The results confirm that our novel MHASRF achieves high recall, ensuring that most no-show patients are correctly identified to reduce missed appointments. In conclusion, based on the validation results, MHASRF demonstrated the most stable and balanced performance. This balance, illustrated in Figure 3, shows that the model can detect most of no-shows (low FN) while maintaining strong precision and generalization. This balanced performance is crucial in healthcare, where minimizing missed no-show prediction is more important than avoiding a few false alerts.

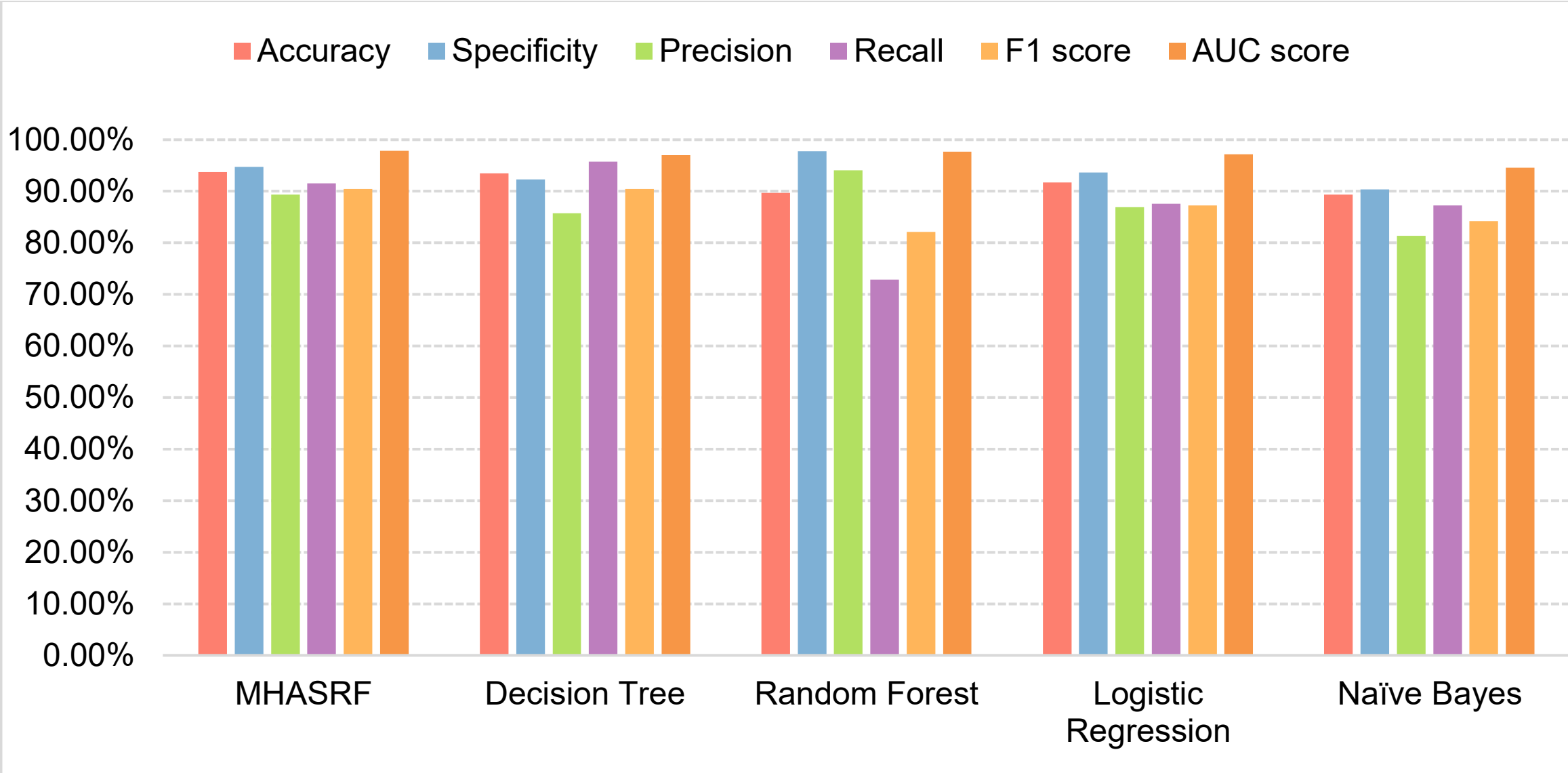

**Figure 3** Model Comparison Results

#### 4.2.2 Loss Curve Analysis

The training and testing loss curves of our proposed model are shown in Figure 4. The training loss consistently decreased over the training epochs, indicating that the model effectively learned from the training data. The testing loss followed a similar trend, demonstrating the model's ability to predict patient no-shows consistently across both training and testing datasets, thereby indicating robust performance.

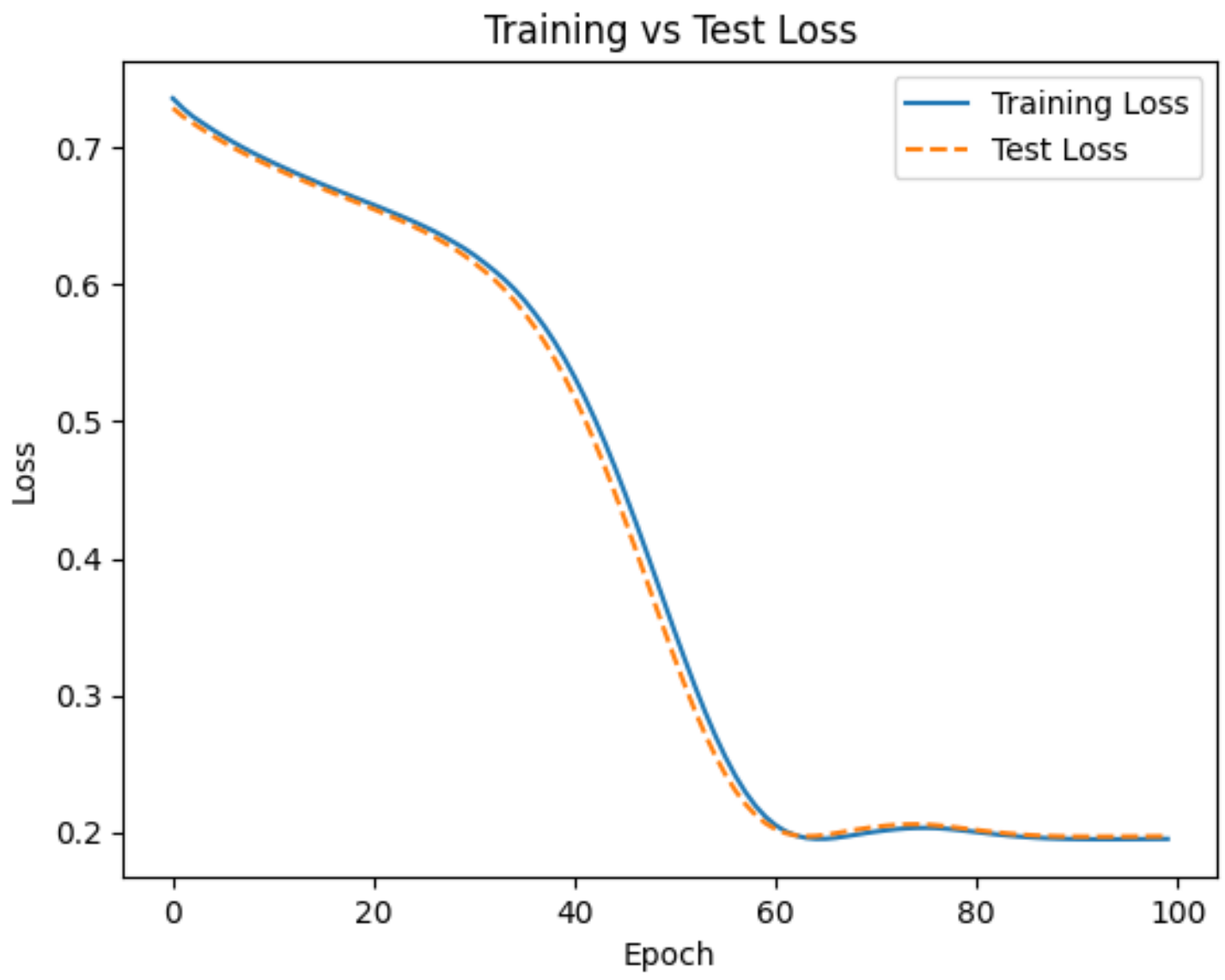

**Figure 4** Training and Testing Loss Curves

### 4.2.3 Attention Weight Distribution

Figure 5 presents an example of the attention weights assigned to each tree in the MHASRF model for a batch of 32 instances. This visualization shows the dynamic behavior of tree attention distribution across different samples.

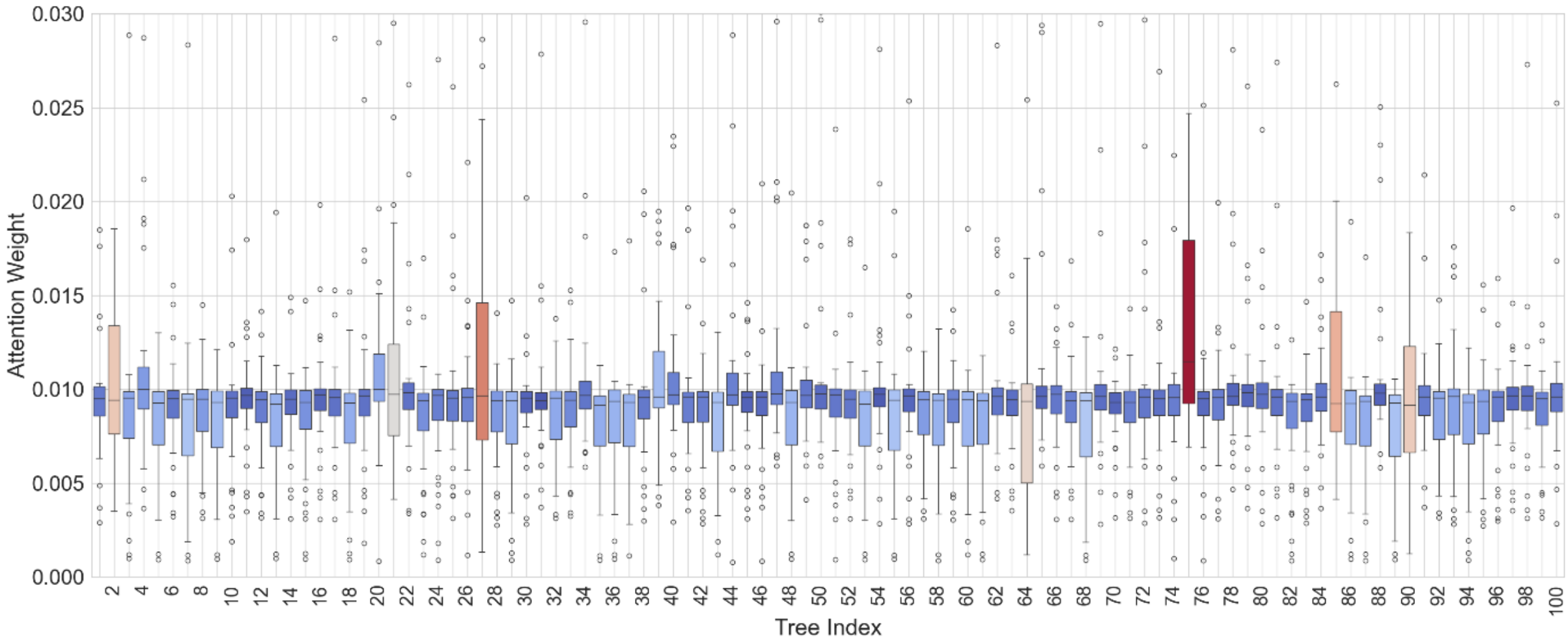

(a)

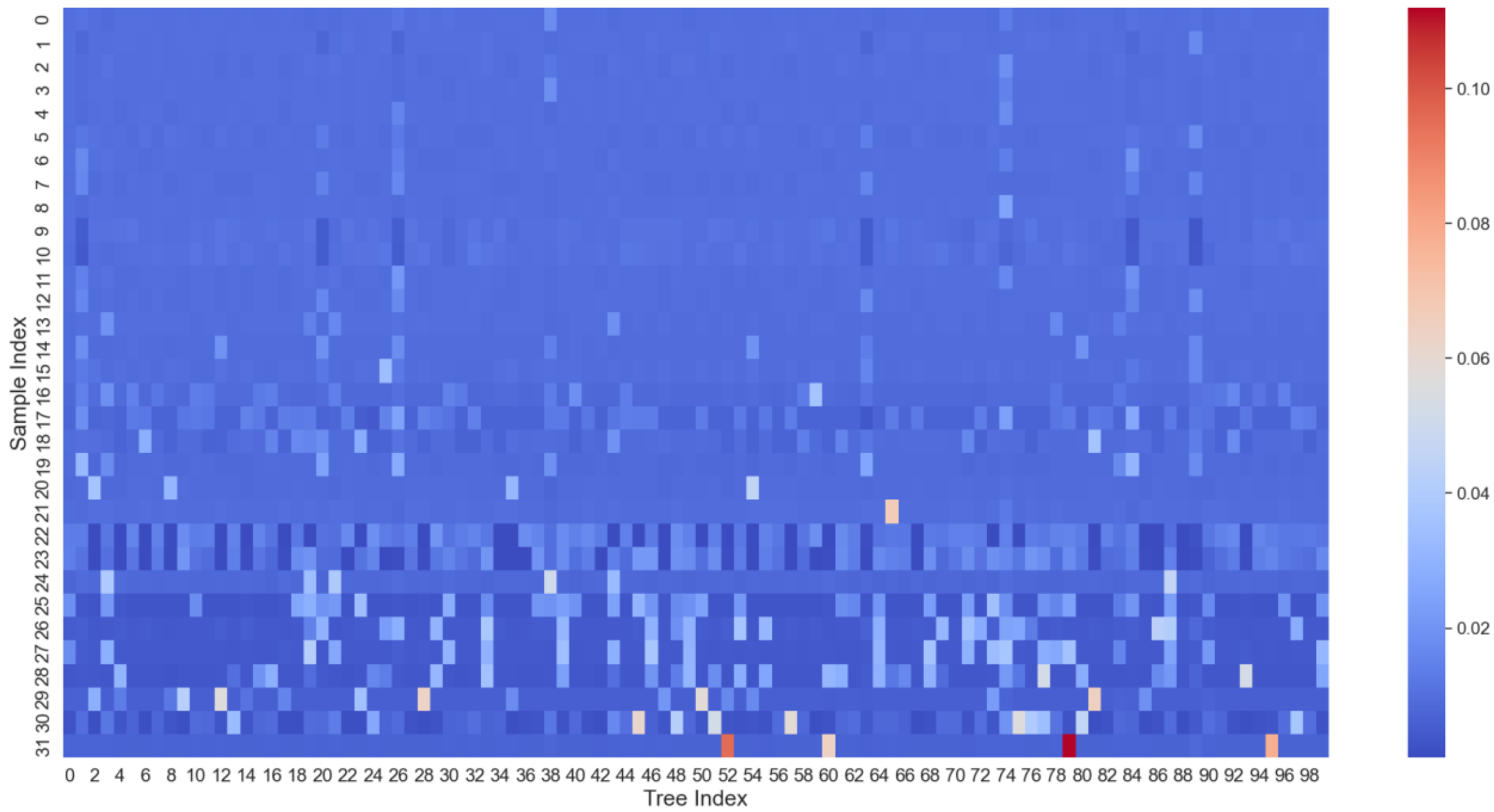

(b)

**Figure 5** Example of Attention Weight Distribution Across Trees for 32 Patient Samples. **a** Box plot showing the distribution of attention weight per tree across samples. **b** Heatmap showing attention weights per tree for each patient.

Figure 5a illustrates the distribution of attention weights for each tree across the 32 samples. The interquartile range (IQR) of each boxplot reveals how consistently each tree contributes across samples. Trees with short boxplots (narrow IQR), such as Tree 11, Tree 31, and Tree 54, exhibit tightly clustered distributions, indicating that their attention weights are consistently distributed across samples. This suggests that these trees contribute similarly in most predictions, likely capturing general patterns of features (e.g., day of the week or patient age), which frequently influence no-show risk uniformly across the population. In contrast, trees with long boxplots (wide IQR), such as Tree 75 and Tree 27, show greater variability in attention weights across samples. These trees contribute differently depending on input characteristics, which implies they respond to more personalized or fluctuating factors (e.g., %No-show or number of past visits).

Figure 5b presents a visualization of how attention is distributed per tree per sample using a heat map with the tree index on the x-axis and the sample index on the y-axis. This figure illustrates a clear patterned contribution, with certain trees highly weighted only for specific subsets of instances and some others not. For instance, Tree 80 might have more weight for certain samples than for others.

These visualizations collectively demonstrate that the attention mechanism in MHASRF allows for instance-specific tree weighting and adapts based on the individual patient behavior. This adaptation makes the model more flexible and suitable for predicting patient no-shows in real healthcare settings, where patient behavior can vary a lot. Instead of giving equal weight across the trees, our model learns to focus on different trees depending on the situation, which helps improve prediction accuracy. Additionally, our instance-specific and adaptive model can capture hidden and unpredictable patterns in patient no-show behavior, allowing better targeted interventions.

#### 4.2.4 Feature Importance

Figure 6 shows feature importance scores obtained from our model. Appointment-related factors such as Visit Reason, %No-show, and Visit Type had the highest impact, highlighting that past behavior is a strong predictor of future no-shows. Among these factors, Visit Reason is the most influential, contributing approximately 26% to the prediction.

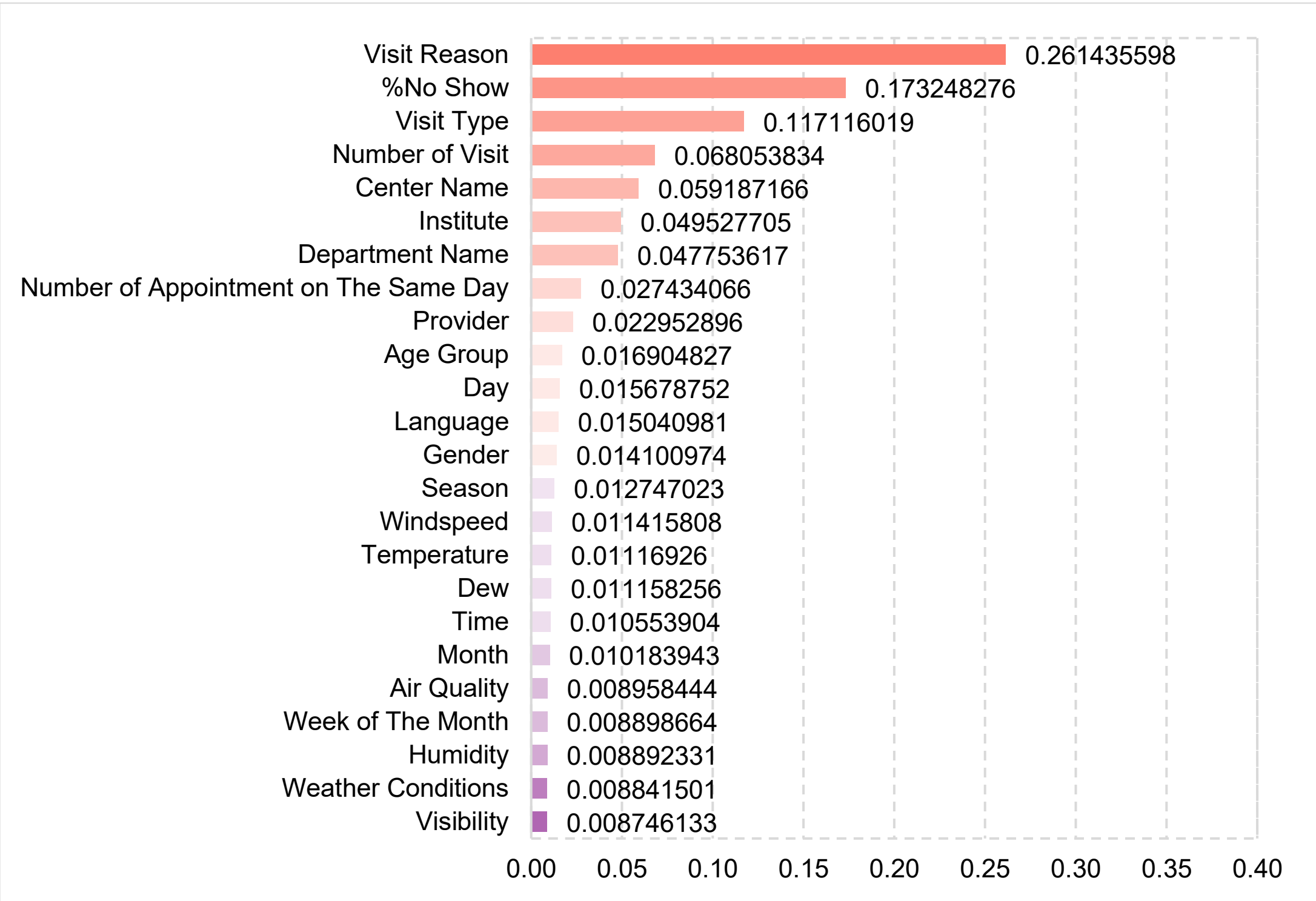

**Figure 6** Feature Importance Scores

These findings indicate that the reason for the appointment plays a significant role in whether patients attend. %No-show, contributing 17.32%, is the second most important feature, reflecting the relevance of no-show history in predicting future absences: patients with a history of missed appointments are more likely to do so again. Visit type, accounting for 11.71%, ranks third in importance, indicating that the nature of the appointment (e.g., new visit or established visit) influences attendance rates.

The results confirm that visit reason, percentage of past no-show history, visit type, and number of previous appointments are four of the most important predictors, which is consistent with findings from previous studies [6][10][26][31]. Our model also identifies additional influential factors that have received less attention in prior research, such as the center name, institute, and department name. These findings suggest that our model can effectively capture feature importance in a manner consistent with other feature selection techniques while also identifying previously underexplored variables.

#### 4.2.5 Ablation Study

An ablation study was conducted to analyze the impact of each model component. The contributions of key components of our model are illustrated in the ablation study results (Table 4). Overall, MHASRF with 3 heads achieved the best performance across all metrics, with an accuracy of 93.72%, specificity of 94.77%, precision of 89.38%, recall of 91.54%, F1 score of 90.44%, and AUC score of 97.87%. The other multi-head configurations produced comparable results. MHASRF (2 heads) obtained 93.38% accuracy, 94.84% specificity, 89.37% precision, 90.34% recall, 89.85% F1 score, and an AUC of 97.61%. Increasing the number of heads to 4 resulted in 93.42% accuracy, 94.20% specificity, 88.38% precision, 91.81% recall, an F1 score of 90.06%, and an AUC of 97.75%. MHASRF with 8 heads achieved 93.50% accuracy, 94.32% specificity, 88.60% precision, 91.81% recall, an F1 score of 90.17%, and an AUC of 97.67%. Removing certain components, such as replacing multi-head attention with single-head attention (SHASRF) or excluding the trainable tree reliability parameter $\delta_k$, led to minor performance degradation. The SHASRF model exhibited slightly lower accuracy (93.36%), specificity (94.55%), recall (90.90%), F1 score (89.89%), and AUC score (97.58%) compared to the best MHASRF configuration. The model without $\delta_k(x)$ achieved the same accuracy (93.36%) as SHASRF but showed lower specificity (94.03%), precision (88.10%), and F1 score (89.99%), and AUC score (97.46%), although it obtained the highest recall (91.96%) among all models.

**Table 4** Ablation Study Results

| Model | Accuracy | Specificity | Precision | Recall | F1 score | AUC score |
|---|---|---|---|---|---|---|
| SHASRF | 93.36% | 94.55% | 88.90% | 90.90% | 89.89% | 97.58% |
| MHASRF (2 heads) | 93.38% | 94.84% | 89.37% | 90.34% | 89.85% | 97.61% |
| MHASRF (3 heads) | 93.72% | 94.77% | 89.38% | 91.54% | 90.44% | 97.87% |
| MHASRF (4 heads) | 93.42% | 94.20% | 88.38% | 91.81% | 90.06% | 97.75% |
| MHASRF (8 heads) | 93.50% | 94.32% | 88.60% | 91.81% | 90.17% | 97.67% |
| MHASRF - without $\delta_k$ | 93.36% | 94.03% | 88.10% | 91.96% | 89.99% | 97.46% |

The tree reliability parameter plays a crucial role in adjusting the attention weights assigned to individual trees. Trees with consistently high classification errors receive lower attention weights, thereby preventing unreliable trees from exerting excessive influence on the final prediction. Unlike traditional RFs that assign the same weight to all trees, our approach actively mitigates the impact of weak and irrelevant trees. Removing the tree reliability parameters resulted in a slight performance decline, with accuracy dropping to 93.36%, confirming that weighting trees based on reliability effectively reduces the influence of poorly performing trees.

Replacing multi-head attention with a single-head resulted in modest reduction in performance; however, MHASRF with three heads consistently achieved higher accuracy, precision, F1 score, and AUC score than SHASRF. Even though the improvements are not large, this is because the model itself already performs very well with AUC near 0.98. The comparison across head numbers also shows that the number of heads mainly changes the balance between precision and recall.

MHASRF with two heads makes the model more conservative, resulting in slightly higher precision and specificity but lower recall, which means that more true no-show cases are missed. In contrast, using four or eight heads increases recall, thereby reducing missed no-shows, but at the cost of lower precision due to an increased number of false positives. The three head-model provides the best balance between these two objectives, achieving the highest F1-score and AUC. In real-world no-show prediction, patients miss appointments for different reasons. By employing multiple attention heads, the model can simultaneously focus on different behavioral patterns. Therefore, even though the performance gains are modest, multi-head attention remains beneficial, as it improves the balance between competing prediction objectives and enhances the model's flexibility for real clinical settings.

## 5. Conclusion

This study successfully developed MHASRF, a hybrid model for predicting patient no-shows, by integrating an attention mechanism and SRF. Our approach used probabilistic soft splits in random forest, multi-head attention, and incorporated trainable tree reliability parameters to improve predictive accuracy, interpretability, instance-specific, and model robustness in comparison to baseline models, such as decision tree, random forest, logistic regression, and naïve Bayes.

MHASRF demonstrated superior performance, achieving high accuracy (93.72), specificity (94.77%), precision (89.38%), recall (91.54%), F1 score (90.44%), and AUC score (97.87%). These results show the effectiveness of integrating attention mechanisms into tree-based models for structured healthcare datasets. Additionally, our model incorporates dynamic tree reliability weights, allowing the model to adjust the influence of each tree based on patient-specific behavioral patterns. This mechanism enhances interpretability and robustness, making MHASRF particularly valuable for clinical decision making and helping medical staff to learn different patterns for patient behavior.

Furthermore, we integrated two levels of feature importance at the tree and attention levels, resulting in our model providing deeper insights into the key factors influencing patient no-shows. The most influential factors contributing to no-show behavior are visit reason, percentage of patient no-show, and visit type. By computing feature importance, our model offers meaningful insight into operational decision-making and appointment management.

Future work can explore expanding this approach to other prediction tasks and incorporating additional factors such as distance from home to hospital, mode of transportation, appointment reminders, and other external factors for predicting patient no-shows that could enhance predictive performance and better capture relationships among features. Moreover, future research could explore subgroup features on each head. Therefore, multi-head attention could capture more diverse patterns and relations that enhance predictive analysis and assist medical staff in capturing more patterns of patient no-show behavior.

### Acknowledgements

This research was supported by Basic Science Research Program through the National Research Foundation of Korea(NRF) funded by the Ministry of Education (No. RS-2023-0024891351382070670001).

### Reference

[1] M. R. Mazaheri Habibi *et al.*, "Evaluation of no-show rate in outpatient clinics with open access scheduling system: A systematic review," Jul. 01, 2024, *John Wiley and Sons Inc*. doi: 10.1002/hsr2.2160.

[2] P. Kheirkhah, Q. Feng, L. M. Travis, S. Tavakoli-Tabasi, and A. Sharafkhaneh, "Prevalence, predictors and economic consequences of no-shows," *BMC Health Serv Res*, vol. 16, no. 1, Jan. 2016, doi: 10.1186/s12913-015-1243-z.

[3] D. L. Nguyen, R. S. DeJesus, and M. L. Wieland, "Missed Appointments in Resident Continuity Clinic: Patient Characteristics and Health Care Outcomes," *J Grad Med Educ*, vol. 3, no. 3, pp. 350–355, Sep. 2011, doi: 10.4300/jgme-d-10-00199.1.

[4] M. J. Drabkin *et al.*, "Telephone reminders reduce no-shows: A quality initiative at a breast imaging center," *Clin Imaging*, vol. 54, pp. 108–111, Mar. 2019, doi: 10.1016/j.clinimag.2018.12.007.

[5] S. Srinivas and A. R. Ravindran, "Optimizing outpatient appointment system using machine learning algorithms and scheduling rules: A prescriptive analytics framework," *Expert Syst Appl*, vol. 102, pp. 245–261, Jul. 2018, doi: 10.1016/j.eswa.2018.02.022.

[6] L. F. Dantas, S. Hamacher, F. L. Cyrino Oliveira, S. D. J. Barbosa, and F. Viegas, "Predicting Patient No-show Behavior: a Study in a Bariatric Clinic," *Obes Surg*, vol. 29, no. 1, pp. 40–47, Jan. 2019, doi: 10.1007/s11695-018-3480-9.

[7] X. Ding *et al.*, "Designing risk prediction models for ambulatory no-shows across different specialties and clinics," *Journal of the American Medical Informatics Association*, vol. 25, no. 8, pp. 924–930, Aug. 2018, doi: 10.1093/jamia/ocy002.

[8] T. H. Almutairi and S. O. Olatunji, "The utilization of AI in healthcare to predict no-shows for dental appointments: A case study conducted in Saudi Arabia," *Inform Med Unlocked*, vol. 46, p. 101472, Jan. 2024, doi: 10.1016/j.imu.2024.101472.

[9] D. Barrera Ferro, S. Brailsford, C. Bravo, and H. Smith, "Improving healthcare access management by predicting patient no-show behaviour," *Decis Support Syst*, vol. 138, Nov. 2020, doi: 10.1016/j.dss.2020.113398.

[10] D. Valero-Bover *et al.*, "Reducing non-attendance in outpatient appointments: predictive model development, validation, and clinical assessment," *BMC Health Serv Res*, vol. 22, no. 1, Dec. 2022, doi: 10.1186/s12913-022-07865-y.

[11] J. Dunstan *et al.*, "Predicting no-show appointments in a pediatric hospital in Chile using machine learning," *Health Care Manag Sci*, vol. 26, no. 2, pp. 313–329, Jun. 2023, doi: 10.1007/s10729-022-09626-z.

[12] A. Alshammari, R. Almalki, and R. Alshammari, "Developing a predictive model of predicting appointment no-show by using machine learning algorithms," *Journal of Advances in Information Technology*, vol. 12, no. 3, pp. 234–239, Aug. 2021, doi: 10.12720/jait.12.3.234-239.

[13] R. Alshammari, T. Daghistani, and A. Alshammari, "The Prediction of Outpatient No-Show Visits by using Deep Neural Network from Large Data," 2020. [Online]. Available: www.ijacsa.thesai.org

[14] D. Liu *et al.*, "Machine learning approaches to predicting no-shows in pediatric medical appointment," *NPJ Digit Med*, vol. 5, no. 1, Dec. 2022, doi: 10.1038/s41746-022-00594-w.

[15] S. Chaudhari, V. Mithal, G. Polatkan, and R. Ramanath, "An Attentive Survey of Attention Models," *ACM Trans Intell Syst Technol*, vol. 12, no. 5, Oct. 2021, doi: 10.1145/3465055.

[16] G. Montavon, W. Samek, and K. R. Müller, "Methods for interpreting and understanding deep neural networks," Feb. 01, 2018, *Elsevier Inc.* doi: 10.1016/j.dsp.2017.10.011.

[17] C. Y. Kuo, L. C. Yu, H. C. Chen, and C. L. Chan, "Comparison of models for the prediction of medical costs of spinal fusion in Taiwan diagnosis-related groups by machine learning algorithms," *Healthc Inform Res*, vol. 24, no. 1, pp. 29–37, Jan. 2018, doi: 10.4258/hir.2018.24.1.29.

[18] J. Li, J. Shi, J. Chen, Z. Du, and L. Huang, "Self-attention random forest for breast cancer image classification," *Front Oncol*, vol. 13, Feb. 2023, doi: 10.3389/fonc.2023.1043463.

[19] L. V. Utkin and A. V. Konstantinov, "Attention-based random forest and contamination model," *Neural Networks*, vol. 154, pp. 346–359, Oct. 2022, doi: 10.1016/j.neunet.2022.07.029.

[20] N. Frosst and G. Hinton, "Distilling a Neural Network Into a Soft Decision Tree," Nov. 2017, [Online]. Available: http://arxiv.org/abs/1711.09784

[21] L. Breiman, "Random Forests," *Machine Learning* , vol. 45, pp. 5–32, 2001, doi: 10.1023/A:1010933404324.

[22] S. AlMuhaideb, O. Alswailem, N. Alsubaie, I. Ferwana, and A. Alnajem, "Prediction of hospital no-show appointments through artificial intelligence algorithms," *Ann Saudi Med*, vol. 39, no. 6, pp. 373–381, 2019, doi: 10.5144/0256-4947.2019.373.

[23] M. U. Ahmad, A. Zhang, and R. Mhaskar, "A predictive model for decreasing clinical no-show rates in a primary care setting," *Int J Healthc Manag*, vol. 14, no. 3, pp. 829–836, 2021, doi: 10.1080/20479700.2019.1698864.

[24] T. Batool, M. Abuelnoor, O. El Boutari, F. Aloul, and A. Sagahyroon, “Predicting Hospital No-Shows Using Machine Learning,” in *IoTaIS 2020 - Proceedings: 2020 IEEE International Conference on Internet of Things and Intelligence Systems*, Institute of Electrical and Electronics Engineers Inc., Jan. 2021, pp. 142–148. doi: 10.1109/IoTaIS50849.2021.9359692.

[25] R. M. Goffman *et al.*, “Modeling patient no-show history and predicting future outpatient appointment behavior in the veterans health administration,” *Mil Med*, vol. 182, no. 5, pp. e1708–e1714, May 2017, doi: 10.7205/MILMED-D-16-00345.

[26] K. Topuz, H. Uner, A. Oztekin, and M. B. Yildirim, “Predicting pediatric clinic no-shows: a decision analytic framework using elastic net and Bayesian belief network,” *Ann Oper Res*, vol. 263, no. 1–2, pp. 479–499, Apr. 2018, doi: 10.1007/s10479-017-2489-0.

[27] I. Mohammadi, H. Wu, A. Turkcan, T. Toscos, and B. N. Doebbeling, “Data Analytics and Modeling for Appointment No-show in Community Health Centers,” *J Prim Care Community Health*, vol. 9, Nov. 2018, doi: 10.1177/2150132718811692.

[28] A. F. A. Hamdan and A. A. Bakar, “Machine Learning Predictions on Outpatient No-Show Appointments in a Malaysia Major Tertiary Hospital,” *Malaysian Journal of Medical Sciences*, vol. 30, no. 5, pp. 169–180, 2023, doi: 10.21315/mjms2023.30.5.14.

[29] E. Abushaaban and M. Agaoglu, “Medical Appointment No-Show Prediction Using Machine Learning Techniques,” in *2022 2nd International Conference on Computing and Machine Intelligence, ICMI 2022 - Proceedings*, Institute of Electrical and Electronics Engineers Inc., 2022. doi: 10.1109/ICMI55296.2022.9873652.

[30] G. Fan, Z. Deng, Q. Ye, and B. Wang, “Machine learning-based prediction models for patients no-show in online outpatient appointments,” *Data Science and Management*, vol. 2, pp. 45–52, Jun. 2021, doi: 10.1016/j.dsm.2021.06.002.

[31] E. Ahmadi, A. Garcia-Arce, D. T. Masel, E. Reich, J. Puckey, and R. Maff, “A metaheuristic-based stacking model for predicting the risk of patient no-show and late cancellation for neurology appointments,” *IISE Trans Healthc Syst Eng*, vol. 9, no. 3, pp. 272–291, Jul. 2019, doi: 10.1080/24725579.2019.1649764.

[32] D. Bahdanau, K. Cho, and Y. Bengio, “Neural Machine Translation by Jointly Learning to Align and Translate,” Sep. 2014, [Online]. Available: http://arxiv.org/abs/1409.0473

[33] E. A. Nadaraya, “On Estimating Regression.” [Online]. Available: https://www.mathnet.ru/eng/tvp356

[34] G. S. Watson, “Indian Statistical Institute SMOOTH REGRESSION ANALYSIS,” 1961. [Online]. Available: http://www.jstor.org/page/info/about/policies/terms.jsp.http://www.jstor.org

[35] A. Vaswani *et al.*, “Attention Is All You Need,” Jun. 2017, [Online]. Available: http://arxiv.org/abs/1706.03762

[36] L. V. Utkin, A. V. Konstantinov, and S. R. Kirpichenko, “Attention and self-attention in random forests,” *Progress in Artificial Intelligence*, vol. 12, no. 3, pp. 257–273, Sep. 2023, doi: 10.1007/s13748-023-00301-0.

[37] D. Zhu, H. Yao, B. Jiang, and P. Yu, “Negative Log Likelihood Ratio Loss for Deep Neural Network Classification,” Apr. 2018, [Online]. Available: http://arxiv.org/abs/1804.10690

[38] N.N. Amalina, “Prediction-driven adaptive double-booking strategy for outpatient scheduling: a multi-objective reinforcement learning approach,” 2025, Master’s thesis, Kumoh National Institute of Technology.